\documentclass[runningheads]{llncs}
\usepackage[T1]{fontenc}
\usepackage{graphicx,verbatim}
\usepackage{epstopdf}
\usepackage{amsmath,amssymb}
\usepackage{booktabs}
\usepackage{multicol,multirow}
\usepackage{hyperref}
\usepackage[dvipsnames]{xcolor} % Required for custom colors
\hypersetup{
    colorlinks=true,        % False: boxed links; True: colored links
    linkcolor=MidnightBlue, % Color of internal links (sections, pages)
    citecolor=MidnightBlue, % Color of bibliographic citations
    filecolor=magenta,      % Color of file links
    urlcolor=MidnightBlue,  % Color of external links
    bookmarksnumbered=true, % Put section numbers in PDFs
    pdfpagemode=FullScreen,% Show bookmarks in PDF viewer
    pdftitle={VAEsselSparse},
}
\usepackage{orcidlink}

\newcommand{\method}{VAEsselSparse}
\begin{document}

\title{Sparse Representation Learning for Vessels}
\titlerunning{VAEsselSparse}

\author{
    Chinmay~Prabhakar \inst{1,2} \orcidlink{0000-0002-1780-8108}\and
    Bastian Wittmann \inst{1,2} \orcidlink{0000-0002-6308-2926} \and
    Paul Büschl \inst{1} \orcidlink{0009-0002-6685-241X} \and 
    Hongwei~Bran~Li\inst{3} \orcidlink{0000-0002-5328-6407}\and
    Bjoern~Menze\inst{1}\thanks{Contributed equally as senior authors} \orcidlink{0000-0003-4136-5690}  \and
    Suprosanna~Shit$^\star$ \inst{1,2} \orcidlink{0000-0003-4435-7207}
  }

\authorrunning{Prabhakar et al.}
\institute{Department of Quantitative Biomedicine, University of Zurich, Switzerland \and
ETH AI Center, ETH Zurich, Switzerland \and
Department of Diagnostic Radiology, National University of Singapore, Singapore
\email{chinmay.prabhakar@uzh.ch}}

\maketitle              % typeset the header of the contribution
\begin{abstract}
Analyzing human vasculature and vessel-like, tubular structures, such as airways, is crucial for disease diagnosis and treatment. Current methods often rely on small sub-regions or simplified tree-like structures, rendering analysis of entire organ-level networks at clinical resolution computationally challenging. To this end, we propose VAEsselSparse, an efficient encoder-decoder model to obtain a meaningful yet compact representation of the entire organ-level vascular network at sub-millimeter resolution. VAEsselSparse leverages the inherent sparsity of 3D vascular structures via sparse convolutions and attention mechanisms, achieving substantial spatial compression rates of $8 \times 8 \times 8$. 
We demonstrate superior reconstruction performance compared to dense counterparts and previous methods. Importantly, the resulting latent space retains clinically relevant discriminative features readily usable for classification tasks, such as aneurysm/stenosis or subvariants of the circle of Willis. Moreover, the compact latent space of VAEsselSparse serves as an effective representation for learning vessel-specific priors through generative models, enabling the synthesis of realistic vasculature.

% Authors must provide keywords and are not allowed to remove this Keyword section.

\end{abstract}
\section{Introduction}
\label{sec:intro}
The human vasculature and other vessel-like anatomical structures, such as airway trees, play essential roles in clinical diagnosis~\cite{shi2020clinically}, risk stratification~\cite{lin2022deep}, and treatment planning~\cite{alhonnoro2010vessel} across numerous cerebrovascular and respiratory conditions. Typically, vessel-like networks are extracted from imaging modalities using deep learning-based segmentation models. 
For many clinical applications, capturing small vessels, often at sub-millimeter resolution, is essential. This, however, especially when segmenting entire organ-level vascular structures, results in 3D segmentation maps of considerable size. Consequently, learning effective representation spaces for high-resolution volumetric data, suitable for generative modeling or classification tasks, poses significant computational challenges.

For generative modeling, existing methods commonly employ simplifying assumptions such as tree-only \cite{feldman2023vesselvae,feldman2025vesselgpt}, spline models \cite{feldman2025vesselgpt}, or metric graphs \cite{prabhakar20243d,prabhakar2025semantically}. While these simplifications reduce computational complexity, they sacrifice essential 3D geometric details \cite{prabhakar20243d,prabhakar2025semantically} or limit analysis to smaller, tree-only regions \cite{alhonnoro2010vessel,feldman2025vesselgpt} of interest. Additionally, the typical patch-based representation-learning strategy \cite{varma2025medvae} used for real-valued 3D signals is suboptimal for organ-level vessel geometries, as they fail to capture the global context of the entire vessel network. To address these limitations, our work aims to develop a computationally efficient method that learns meaningful representations of organ-level vascular networks at sub-millimeter 3D resolution, a critical gap in existing methods.

Unlike real-valued 3D images, vascular segmentation maps are inherently sparse, motivating the need for sparsity-preserving encoder-decoder architectures for efficient representation learning. Additionally, it is essential to preserve coarse, global vessel structure within the latent space and prevent fine structural losses due to averaging effects from dense convolution. To address these requirements, we propose an efficient method that leverages sparse convolution and sparse transformer-based encoder-decoder architectures. Our approach efficiently manages the computational demands of encoding large-scale 3D vessel networks at the organ level, achieving a high spatial compression rate ($8\times8\times8$) while maintaining a structured latent space ideal for classification and generative modeling tasks. In summary, our contribution is as follows:

\begin{enumerate}
\item We propose \method, an efficient encoder-decoder architecture designed for learning compact representations of large-scale, organ-level 3D vascular networks, even at sub-millimeter resolution. \footnote{\href{https://github.com/chinmay5/sparse_representation_learning_for_vessels/tree/main}{https://github.com/chinmay5/sparse\_representation\_learning\_for\_vessels/}}
\item \method{} achieves superior reconstruction performance compared to dense architectures and significantly surpasses previous volumetric VAEs in terms of compression rates while maintaining structural connectivity.
\item Our learned latent representation effectively supports downstream applications, including vascular classification tasks and generative modeling.
\end{enumerate}

\section{Related Literature}
\label{sec:rel_lit}

\textbf{Vessel Representations:} Early works on vascular modeling were mostly GAN-based. Wolternik et al.~\cite{wolterink2018blood}, for example, relied on GANs to synthesize coronary artery geometries. VesselVAE \cite{feldman2023vesselvae} models vascular meshes using an autoregressive model. VesselVAE, however, is limited to tree-only shapes with a maximum node degree of three. In contrast, two-stage discrete diffusion models \cite{prabhakar20243d,prabhakar2025semantically} are capable of generating arbitrary networks and have been explored on simplified metric graphs of vessels and airways. Recently, multiple part-based models \cite{chen2025hierarchical,batten2025vector} have been proposed to encode individual branches stitched by a branch configuration mechanism. Further prior works explored the use of signed-distance fields \cite{kuipers2024generating,kuipers2025self} to learn the implicit shapes of small field-of-view vasculature. Feldman et al.~\cite{feldman2025vesselgpt} explored transformer-based autoregressive generation of vessel shapes, but this approach is limited to tree-only structures and simplistic shapes.

\noindent
In contrast, our work operates directly on the entire organ-level high-resolution segmentation map and does not rely on \textit{any} simplifying assumptions, thereby improving the clinical utility of the learned representation.

\vspace{0.5em}
\noindent\textbf{3D Sparse Representations:} In computer vision, encoding 3D shapes is a long standing problem. Recently, sparse variants of convolutional~\cite{ren2024xcube,xiang2025structured} and transformer~\cite{wu2025direct3d} architectures have been successfully applied to image-conditioned 3D shape synthesis. These works improve both the efficiency of 3D shape processing and the fidelity of fine structural details. However, they often train independent hierarchical models at different resolutions for compression, rather than learning a single compact latent. For example, \cite{ren2024xcube} proposed a hierarchical encoding model for coarse-to-fine generation and refinement, scalable up to $1024^3$.

\noindent
In this work, we, for the first time, re-purpose and unify these concepts to sparse 3D vascular structures, resulting in expressive latents of superior compression rates, maintaining fine-grained details of sub-millimeter structures.

\section{\method}
\label{sec:method}

\begin{figure}[t!]
    \centering
    \includegraphics[width=\textwidth, trim=27 1 20 15, clip]{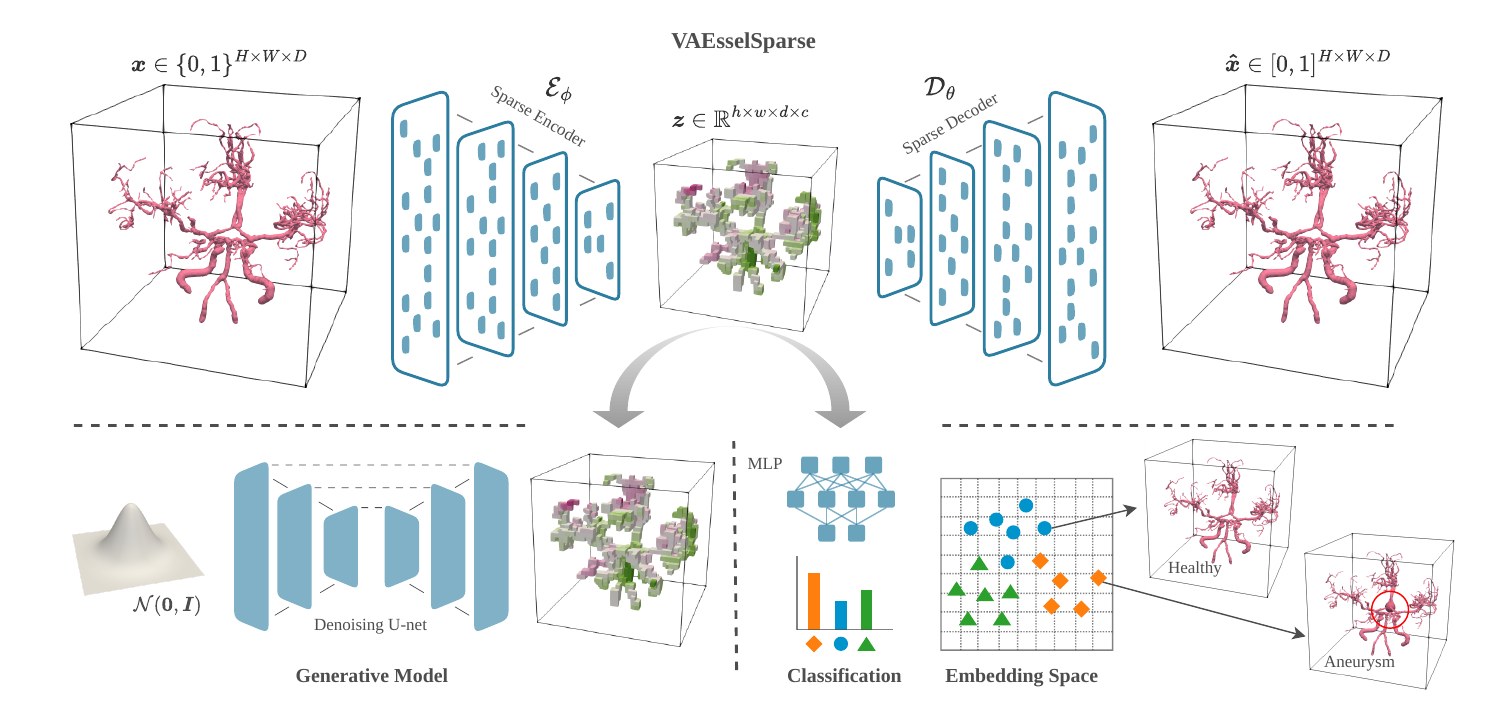}
    \caption{Overview. \method{} represents an encoder-decoder architecture comprised of sparse convolution and sparse attention layers and is, therefore, tailored to operate on sparse input masks. Our learned latent space is suitable for a wide variety of tasks, inclding generative modeling and classification.
    }
    \label{fig:overview}
\end{figure}

Our objective is to learn an encoder-decoder model to compress a sparse volumetric vessel segmentaion map $\boldsymbol{x}\in \{0,1\}^{H\times W\times D}$, where $\|\boldsymbol{x}\|_{\ell_0}\ll H \cdot W \cdot D$, into a latent of size $\boldsymbol{z} \in \mathbb{R}^{h\times w\times d \times c}$. Here, $H, W, D$ is defined as the original voxel spatial dimension, $h, w, d$ as the latent space spatial dimension, and $c$ represents the number of features in our latent space. 
We subsequently define a \emph{spatial compression ratio} ($r_s=\frac{H}{h}=\frac{W}{w}=\frac{D}{d}$), and a \emph{volumetric compression ratio} ($r=\frac{H\cdot W\cdot D}{h\cdot w\cdot d\cdot c}$). Since we want to preserve the sparse nature of the vessel's structure in the latent space, we seek $\|\boldsymbol{z}\|_{\ell_0}\ll h\cdot w\cdot d\cdot c.$ To this end, we adopt sparse-convolution ~\cite{tangandyang2023torchsparse} and sparse attention~\cite{wu2025direct3d} mechanisms to design the \method{}'s architecture as shown in Fig.~\ref{fig:overview}.
Instead of processing a dense $H\times W\times D$ grid, we represent the segmentation as a sparse set of active voxel coordinates
$\mathcal{C}_{\boldsymbol{x}}=\{\boldsymbol{p}\in\mathbb{Z}^3\mid \boldsymbol{x}(\boldsymbol{p})=1\}$.
Each active voxel is associated with a feature vector $(\boldsymbol{F}_{\boldsymbol{x}})$,
forming a sparse tensor $\mathcal{X}=(\mathcal{C}_{\boldsymbol{x}},\boldsymbol{F}_{\boldsymbol{x}})$. We set $\mathcal{X}^{(0)}=\mathcal{X}$.
All sparse convolution and sparse attention operations are applied on $\mathcal{X}$, making the computation scale with $|\mathcal{C}_{\boldsymbol{x}}|$ rather than $H\cdot W\cdot D$.

\vspace{0.5em}
\noindent\textbf{Encoder ($\mathcal{E}_\phi$):}
$\mathcal{E}_\phi$ maps the sparse vessel volume to the parameters of the approximate posterior:
$
  (\boldsymbol{\mu}, \log\boldsymbol{\sigma}^2) = \mathcal{E}_\phi(\boldsymbol{x}),~
  q_\phi(\boldsymbol{z}\mid \boldsymbol{x})=\mathcal{N}(\boldsymbol{\mu},\mathrm{diag}(\boldsymbol{\sigma}^2)).
$ We first extract local geometry using a stack of residual sparse 3D CNN blocks, where each block is composed of sparsity-preserving 3D convolutions, group normalization, and ReLU. This stage aggregates local features along thin structures while avoiding dense 3D computations. To reach the target latent resolution $(h, w, d)$, we downsample the sparse grid using strided sparse convolutions:
\begin{equation}
  \tilde{\mathcal{X}}^{(\ell+1)} = \mathrm{ResSparseBlock}^{(\ell)}\!\left(\mathcal{X}^{(\ell)}\right), \qquad  \mathcal{X}^{(\ell+1)} = \mathrm{Down}^{(\ell)}\!\left(\tilde{\mathcal{X}}^{(\ell+1)}\right), 
\end{equation}
reducing the spatial resolution by a factor of $2$ per stage until the overall factor matches $r_s$.
This yields a bottleneck sparse tensor $\mathcal{X}^{(L)}=(\mathcal{C}_{\boldsymbol{z}},\boldsymbol{F}_{\boldsymbol{z}})$.

Purely convolutional encoders can struggle to preserve global connectivity when aggressively compressing thin structures.
To inject global and mid-range context, we interpret the bottleneck voxels as a set of tokens and add a 3D coordinate-based positional encoding:
$
  \tilde{\boldsymbol{t}}_n = \boldsymbol{F}_{\boldsymbol{z}}(\boldsymbol{p}_n) + \mathrm{PE}(\boldsymbol{p}_n),
$
for $\boldsymbol{p}_n\in\mathcal{C}_{\boldsymbol{z}}$ and process the resulting token set with sparse  windowed self-attention layers~\cite{wu2025direct3d}:
\begin{equation}
  \{\boldsymbol{t}_n\} \leftarrow \mathrm{SparseAttnBlock}\!\left(\{\tilde{\boldsymbol{t}}_n\}\right).
\end{equation}
Our design enables information flow across distant vessel segments while keeping attention costs tractable by restricting interactions to local windows ~\cite{wu2025direct3d}. We sample $\boldsymbol{z}$ via the reparameterization trick:
$
  \boldsymbol{z} = \boldsymbol{\mu} + \boldsymbol{\sigma}\odot\boldsymbol{\epsilon},
  ~
  \boldsymbol{\epsilon}\sim\mathcal{N}(\boldsymbol{0},\boldsymbol{I}).
$, where $\boldsymbol{\mu}$ and $\boldsymbol{\sigma}$ is obtained using two linear layers on $\{\boldsymbol{t}_n\}$.
The latent tensor $\boldsymbol{z}$ is stored in sparse format on the bottleneck coordinates $\mathcal{C}_{\boldsymbol{z}}$.

\vspace{0.5em}
\noindent
\textbf{Decoder ($\mathcal{D}_\theta$):}
$\mathcal{D}_\theta$ mirrors the encoder and reconstructs the vessel volume from $\boldsymbol{z}$ as 
$  \hat{\boldsymbol{x}} = \mathcal{D}_\theta(\boldsymbol{z})$. We first apply sparse transformer blocks to refine the latent tokens with context before spatial upsampling. Let $\tilde{\boldsymbol{u}}_n=\boldsymbol{z}(\boldsymbol{p}_n)+\mathrm{PE}(\boldsymbol{p}_n)$:
\begin{equation}
  \{\boldsymbol{u}_n\} \leftarrow \mathrm{SparseAttnBlock}\!\left(\{\tilde{\boldsymbol{u}}_n\}\right),
\end{equation}
Define the decoder bottleneck tensor $\mathcal{Y}^{(L)}=(\mathcal{C}_{\boldsymbol{z}},\boldsymbol{F}_Y)$ with $\boldsymbol{F}_Y(\boldsymbol{p}_n)=\boldsymbol{u}_n$.
We then progressively upsample the sparse representation using transposed sparse convolutions, followed by refinement via additional sparse convolutional layers:
\begin{equation}
  \tilde{\mathcal{Y}}^{(\ell-1)} = \mathrm{Up}^{(\ell)}\!\left(\mathcal{Y}^{(\ell)}\right),
  \qquad
  \mathcal{Y}^{(\ell-1)} \leftarrow \mathrm{ResSparseBlock}^{(\ell-1)}\!\left(\tilde{\mathcal{Y}}^{(\ell-1)}\right).
\end{equation}

At full resolution, a final $1\times1\times1$ sparse convolutional operation maps features to occupancy logits, followed by a sigmoid activation yielding a probabilistic reconstruction $\hat{\boldsymbol{x}}\in[0,1]^{H\times W\times D}$.

\vspace{0.5em}
\noindent
\textbf{Training Loss:}
We train \method{} with the standard VAE objective comprising a reconstruction term and a KL regularizer:
\begin{equation}
\mathcal{L} = \mathcal{L}_{\mathrm{rec}}(\boldsymbol{x},\hat{\boldsymbol{x}}) \;+\;
\beta\,\mathrm{KL}\!\left(q_\phi(\boldsymbol{z}\mid \boldsymbol{x}) \,\|\, p(\boldsymbol{z})\right),
\end{equation}
where $\beta$ controls the strength of latent regularization. To avoid dense supervision over the full $H\times W\times D$ grid, we compute $\mathcal{L}_{\mathrm{rec}}$ only on the union of active voxels from the ground truth and the reconstruction support. Let $\Omega = \mathcal{C}_{\boldsymbol{x}} \cup \mathcal{C}_{\hat{\boldsymbol{x}}}$ denote this set of voxel coordinates, where $\mathcal{C}_{\hat{\boldsymbol{x}}}$ are the coordinates evaluated by the sparse decoder. This defines $\mathcal{L}_{\mathrm{rec}}$ as:
\begin{equation}
\mathcal{L}_{\mathrm{rec}}
= -\sum_{\boldsymbol{p}\in\Omega}
\Big(
\boldsymbol{x}(\boldsymbol{p})\log \hat{\boldsymbol{x}}(\boldsymbol{p})
+
(1-\boldsymbol{x}(\boldsymbol{p}))\log(1-\hat{\boldsymbol{x}}(\boldsymbol{p}))
\Big).
\end{equation}
\section{Experiments \& Results}
\label{sec:exp}
% In this section, we demonstrate \method{}'s performance on a wide variety of tasks, including reconstruction and downstream classification, and unconditional generative modeling.

\vspace{0.5em}
\noindent\textbf{Downstream Tasks:} To demonstrate \method's efficacy beyond the reconstruction task, we perform two downstream experiments, classification and generative modeling in the \method's learned latent space. For the classification task, we use two variants: (i) Principal Component Analysis (PCA) followed by Random Forest (RF), (ii) an MLP. In both cases, we perform a 3-fold cross-validation. For generative modeling, we train an unconditional denoising U-Net~\cite{guo2025maisi} in the latent space with a flow-matching loss~\cite{lipman2022flow}.

\vspace{0.5em}
\noindent\textbf{Datasets:} We use the following datasets to train, validate, and evaluate the reconstruction performance of \method: (1) ATM~\cite{zhang2023multi}, AIIB~\cite{nan2024hunting}, and AeroPath~\cite{stoverud2023aeropath} for airway trees; (2) COSTA~\cite{mou2024costa} for cerebral vasculature; and (3) HiPas~\cite{chu2025deep}, PARSE~\cite{luo2023efficient}, and Pulmonary-AV~\cite{cheng2024fusion} for pulmonary vessels. We combine these individual datasets into a \emph{single} consolidated dataset at isotropic resolution of 0.5 mm using center-pad, yielding a $640\times 640\times 832$-dimensional volumes, and split into train, validation and test partitions in 80-10-10 ratio, yielding 971 training, 122 validation, and 126 test samples. For the healthy/stenosis/aneurysm classification task, we utilize 160 samples from the INSTED \cite{insted2024} challenge dataset, which is not part of our \method's training data, and extract segmentation maps utilizing the $zero$-shot capable universal foundation model vesselFM \cite{wittmann2025vesselfm}. As a second classification task, we use 126 binary vessel segmentation maps from the TopCoW challenge \cite{yang2025benchmarking} for 8-subtype classification of the circle of Willis. For generative modeling, we use the ATM dataset, consisting of 234 samples.

\vspace{0.5em}
\noindent\textbf{Baselines:}
For reconstruction performance, we compare \method{} against a dense VAE~\cite{guo2025maisi} that has a similar number of parameters and an identical compression ratio. However, due to the dense convolutional model's large GPU VRAM requirements, the dense VAE is trained with a reduced patch size of $96^3$. We utilize a state-of-the-art tensor-split base implementation \cite{guo2025maisi} to train the dense VAE. Further, we compare against the recently proposed MedVAE \cite{varma2025medvae}, which offers a reduced $4\times 4\times 4$ spatial compression ratio, compared to ours of $8\times 8\times 8$. Moreover, we compare \method{} against VesselGPT \cite{feldman2025vesselgpt} on the ATM dataset, as VesselGPT is only applicable to tree-like structures. For the classification task, we compare against the ResNet and ViT baselines operating in the voxel space and against the PCA+RF model on the dense VAE features.

\begin{table}[!t]
\centering
\scriptsize
\caption{Quantitative comparison of our method against dense convolutional baselines. \method~accurately preserves topology while operating at a higher spatial compression ratio, providing a high-fidelity, computationally efficient latent space.}
\label{tab:metrics}
\setlength{\tabcolsep}{6pt}
\begin{tabular}{l|c |c |c |c |c |c}
\toprule
\textbf{Method} & \textbf{$r_s$} $\uparrow$ & \textbf{\# Params (M)}& \textbf{Dice} $\uparrow$ & \textbf{clDice} $\uparrow$ & $|\Delta\beta_0|$ $\downarrow$ & $|\Delta\beta_1|$ $\downarrow$\\
\midrule
Dense VAE & $\mathbf{8\times 8\times 8}$ & 20.10 & 74.79 & 73.48 & 115.79 & 34.61 \\
MedVAE \cite{varma2025medvae} & $4\times 4\times 4$ & 161.00 & \textbf{85.10} & 85.76 & 6.70   & 54.21 \\
\midrule
\method      & $\mathbf{8\times 8\times 8}$ & \textbf{20.09} & 81.45 & \textbf{92.11} & \textbf{5.11}   & \textbf{21.80} \\
\bottomrule
\end{tabular}
\end{table}

\begin{figure}[!t]
    \centering
    \hspace{2em} Reference \hfill \hspace{0.75em} Dense VAE \hfill \hspace{0.75em} MedVAE \cite{varma2025medvae} \hfill \method~(ours) \hfill\\
    \includegraphics[width=\textwidth, trim=0 120 0 70, clip]{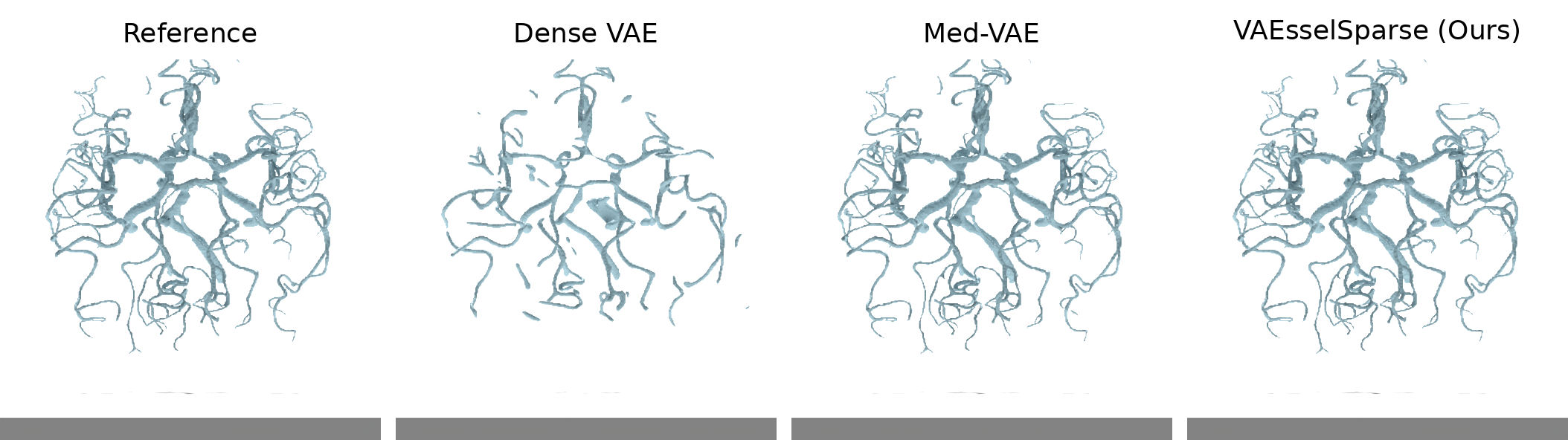}
    \includegraphics[width=\textwidth, trim=0 95 0 150, clip]{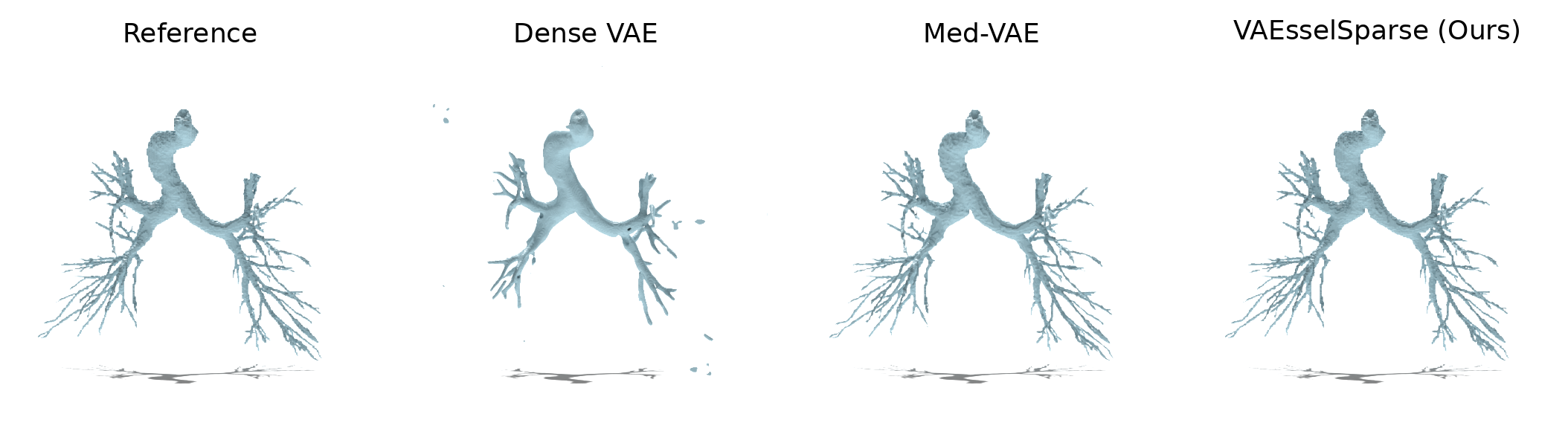}
    \includegraphics[width=\textwidth, trim=0 30 0 190, clip]{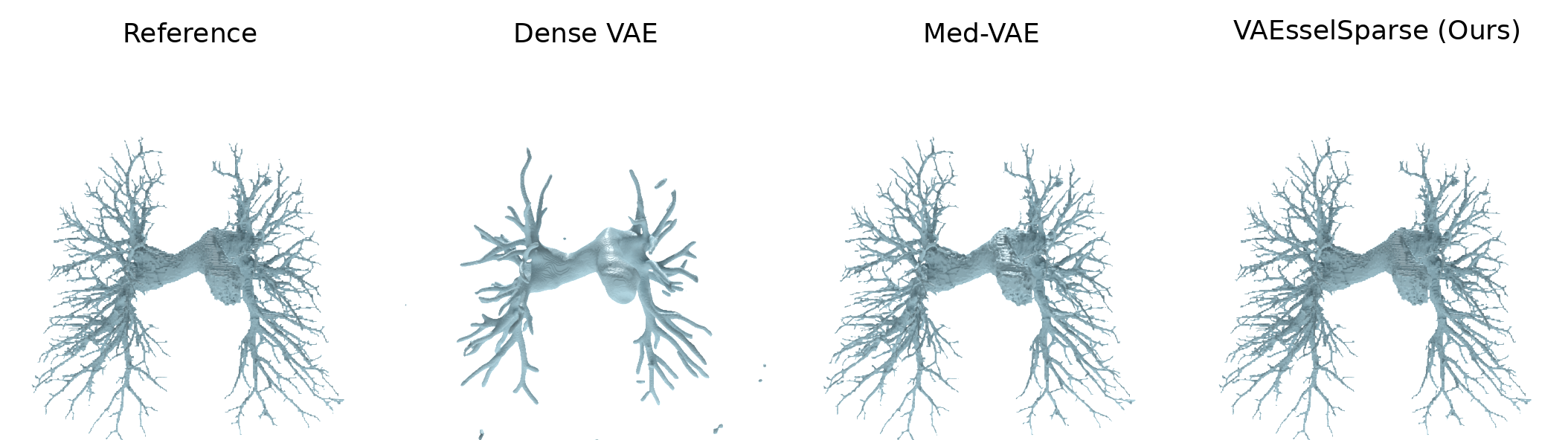}
    \caption{Representative samples from the reconstruction experiments on COSTA (top row), AIB (middle row), and PARSE (bottom row) test sets.}
    \label{fig:recon}
\end{figure}

\vspace{0.5em}
\noindent\textbf{Implementation Details:} \method's encoder consists of three residual downsampling blocks, while the decoder uses residual sparse transpose convolution blocks for upsampling. We employ six sparse transformer blocks, each with eight heads and 384 channels. The dense VAE consists of three encoder-decoder blocks~\cite{guo2025maisi}. We use $c=2$ for both models and for 1000 epochs using the AdamW optimizer with a learning rate of 1e-4. \method~takes approximately 36 hours to train on a single RTX A6000 GPU, while the dense VAE converges in 90 hours. For PCA, we use the first 15 components, and use 200 trees for RF classification. The MLP represents a 3-layer model with 64 hidden dimensions and is trained for 200 epochs with a learning rate of 3e-4. For the generative model, we use a denoising U-Net~\cite{guo2025maisi} with three encoder and decoder blocks and train for 2000 epochs with a learning rate of 1e-5 for both dense VAE and \method.

\begin{table}[!t]
\centering
\scriptsize
\caption{Classification results on INSTED and TopCoW. Results demonstrate that \method{}'s latent
space retains clinically relevant, discriminative features.}
\label{tab:cls_insted_topcow}
\setlength{\tabcolsep}{6pt}
\begin{tabular}{l|cc|cc}
\toprule
\multirow{2}{*}{\textbf{Method}} & \multicolumn{2}{c|}{\textbf{INSTED}} & \multicolumn{2}{c}{\textbf{TopCOW}} \\
\cline{2-5}
 & Bal. Accuracy & Macro-F1 & Bal. Accuracy & Macro-F1\\
\midrule

Resnet & 85.24 & 84.93 & 78.95 & 86.34 \\
ViT   & 73.08 & 55.24 & 80.64 & 86.59 \\
Dense VAE + PCA + RF          & 34.89 & 32.65 & 75.58 & 71.29 \\
\midrule
\method~ + PCA + RF   & 85.93 & 85.31 & 78.01 & 87.68 \\
\method~ + MLP & \textbf{87.96} & \textbf{86.94} & \textbf{81.05} & \textbf{86.84} \\
\bottomrule
\end{tabular}
\end{table}

\begin{figure}[!t]
    \centering
    \includegraphics[width=0.45\textwidth]{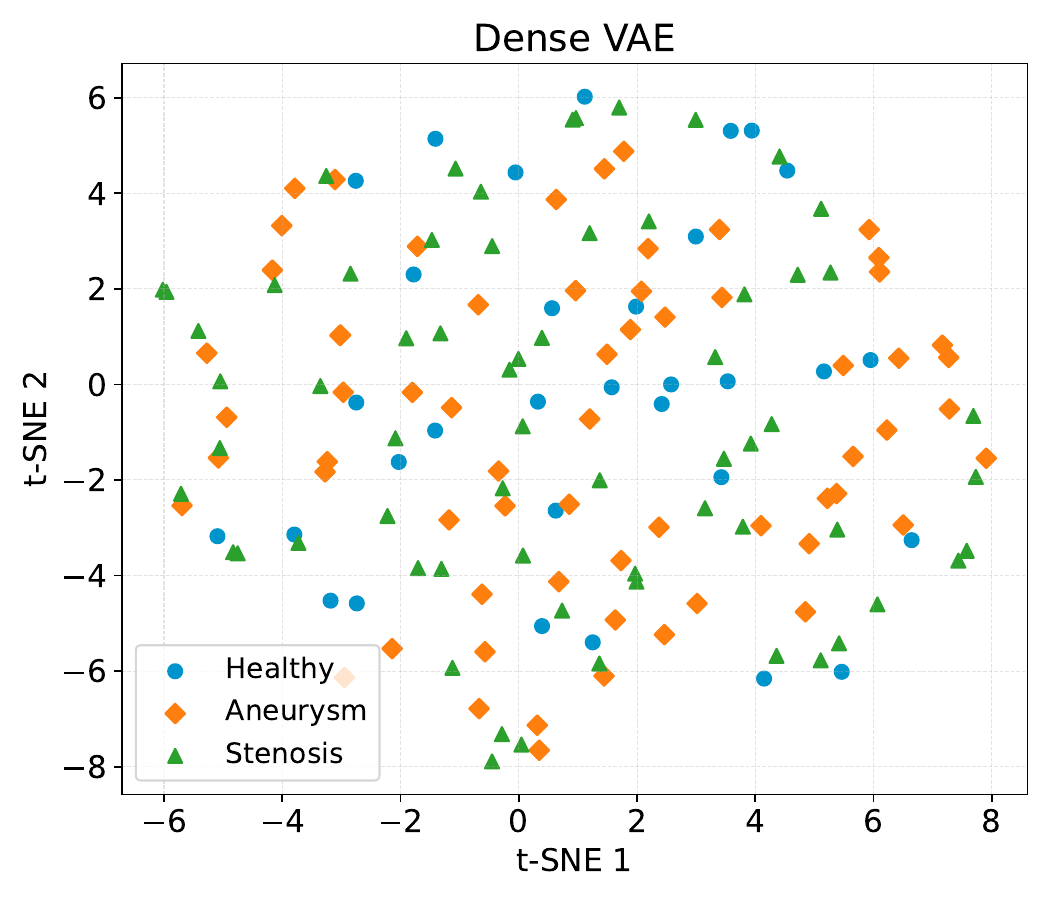}
    \includegraphics[width=0.45\textwidth]{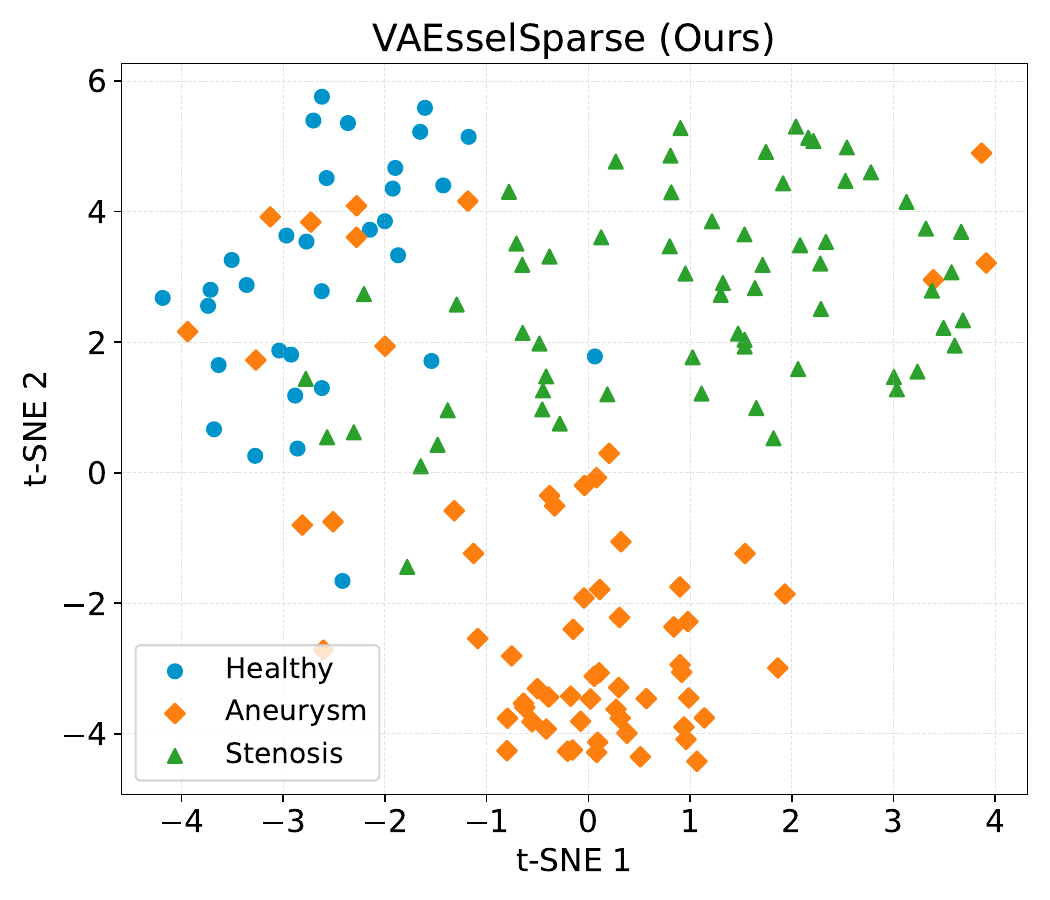}
    \caption{t-SNE visualization of 15 PCA components of \method's latent space on the INSTED dataset. The analysis reveals emerging clusters based on the classes healthy/stenosis/aneurysm (right), whereas the dense VAE fails to capture any (left).}
    \label{fig:tsne}
\end{figure}

\vspace{0.5em}
\noindent\textbf{Metrics:}
For the reconstruction task, we report Dice, clDice \cite{shit2021cldice}, mean absolute Betti 0 error ($|\Delta\beta_0|$), and Betti 1 error ($|\Delta\beta_1|$). For the classification task, we report the balanced accuracy and mean F1 scores, while for the generation task, we report Minimum Matching Distance (MMD), Coverage (COV), and 1-Nearest Neighbor Accuracy (1-NNA) following \cite{feldman2025vesselgpt}.

\vspace{0.5em}
\noindent\textbf{Reconstruction Results:}
Table~\ref{tab:metrics} shows that \method~outperforms the dense VAE baseline on all metrics. Further, we observe that it achieves a comparable Dice score to MedVAE, which offers significantly less spatial compression and consists of significantly more parameters. Qualitatively, Fig. \ref{fig:recon} mirrors our quantitative findings and demonstrates that \method~preserves connectivity better compared to the dense VAE. Note that, unlike MedVAE, we did not need the GAN loss during training. To compare to vessel-specific methods, we compare \method~against VesselGPT \cite{feldman2025vesselgpt} on the ATM dataset. We observe that VesselGPT struggles with complex vessel structures and only achieves a clDice of 1.64, where MedVAE achieves 84.61, while \method~outperforms all methods with, achieving a clDice of 90.09.

\vspace{0.5em}
\noindent\textbf{Classification Results:}
We show that \method's latent space contains vessel-specific discriminative features, yielding superior classification results on both the INSTED and TopCoW datasets, which were unseen to \method~during training, with a simple MLP over strong ResNet and ViT baselines. Even with PCA+RF, \method~achieves competitive performance, whereas the dense VAE model underperforms. We observe that ViT struggles on the INSTED dataset, where it severely overfits. Further, to visualize the learned representation, we examine a t-SNE plot of the top 15 PCA features in the latent space, which reveals separate clusters for the healthy/aneurysm/stenosis categories in our method, whereas dense VAE shows no such clusters (Fig.~\ref{fig:tsne}).

\begin{figure}[!t]
    \centering
     \includegraphics[width=0.2\textwidth, trim=160 320 160 10, clip]{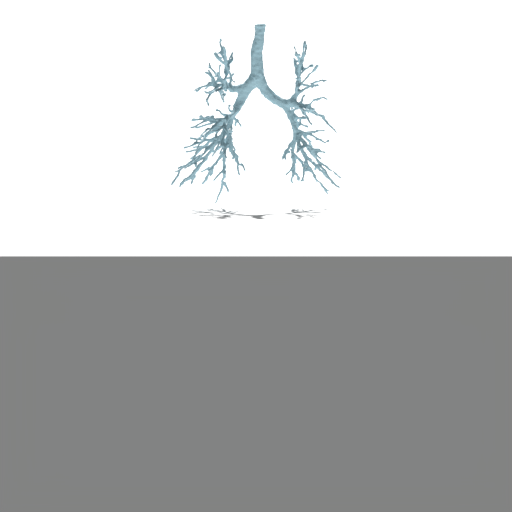}
     \includegraphics[width=0.2\textwidth, trim=160 320 160 10, clip]{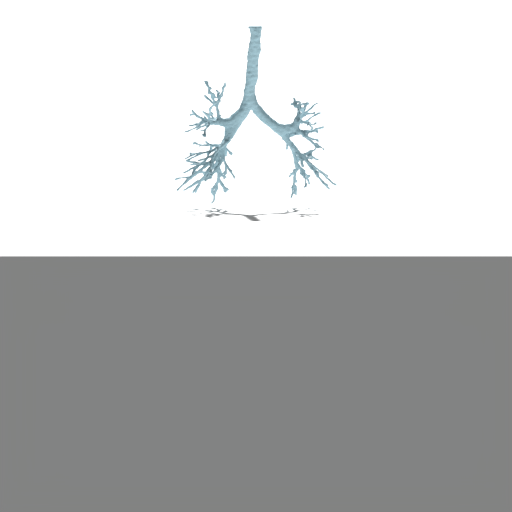}
     \includegraphics[width=0.2\textwidth, trim=160 320 160 10, clip]{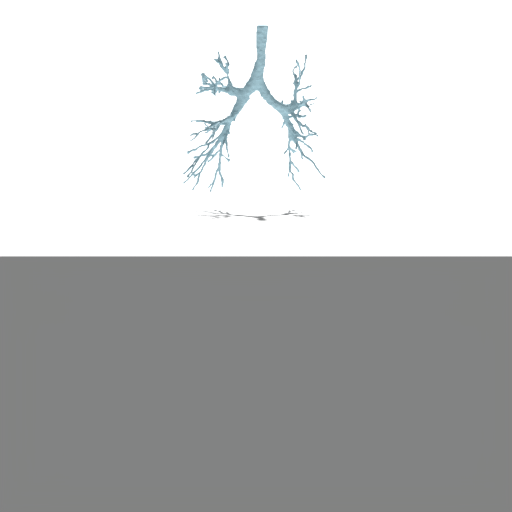}
     \includegraphics[width=0.2\textwidth, trim=160 320 160 10, clip]{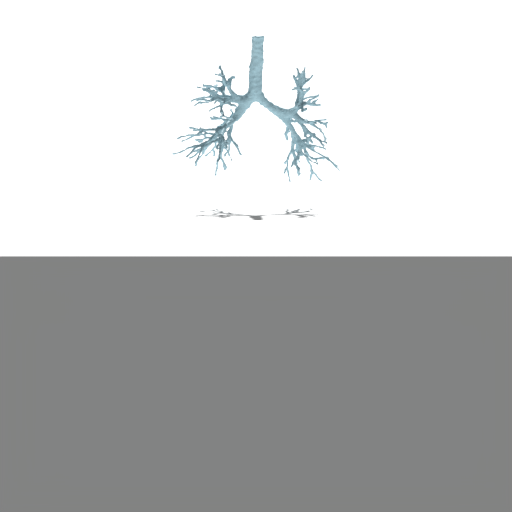}\\
     \includegraphics[width=0.2\textwidth, trim=160 320 160 10, clip]{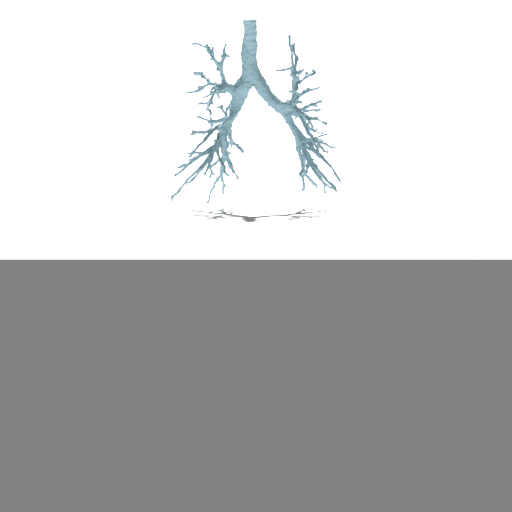}
     \includegraphics[width=0.2\textwidth, trim=160 320 160 10, clip]{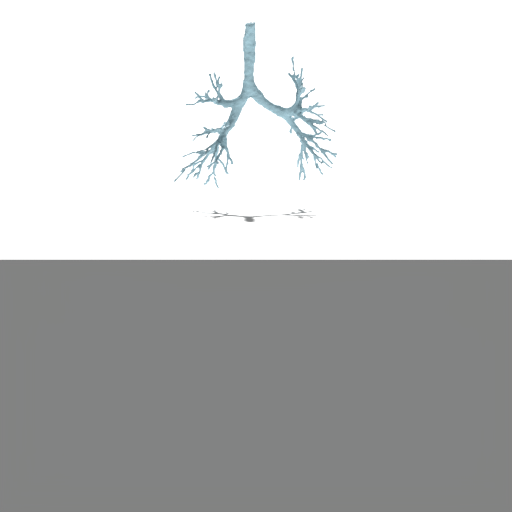}
     \includegraphics[width=0.2\textwidth, trim=160 320 160 10, clip]{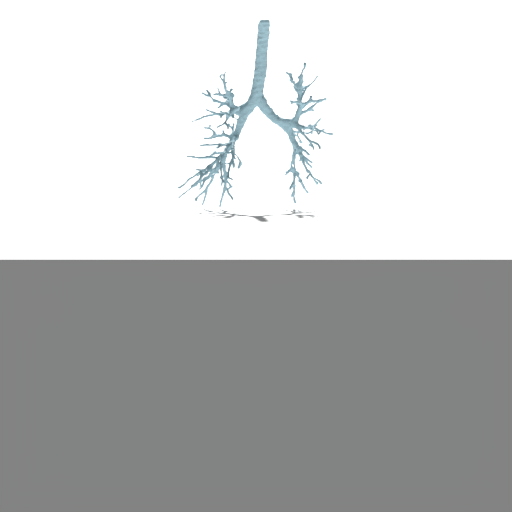}
     \includegraphics[width=0.2\textwidth, trim=160 320 160 10, clip]{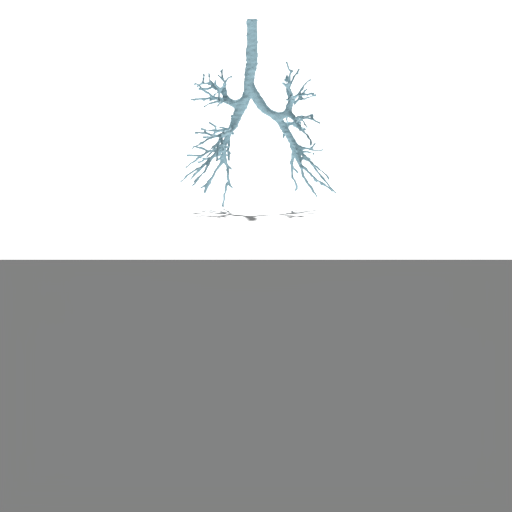}
    \caption{Unconditional generated samples from the denoising U-Net-based flow-matching model trained on \method's latent space of the ATM dataset.}
    \label{fig:genration}
\end{figure}

\vspace{0.5em}
\noindent\textbf{Generation Results:}
We find that under identical settings, a denoising U-Net produces more realistic vessels when trained with the \method's latent space over the dense VAE. We report MMD of \textbf{25.62} for \method{} vs. 63.41 for dense VAE. Similarly, a COV value of \textbf{46.87} vs. 38.75, and 1-NN of \textbf{60.16} vs. 85.93. Fig. \ref{fig:genration} shows typical volumes generated by \method. We observe that the generative model struggles in dense VAE latent space due to its high dimensionality relative to the limited sample size, whereas \method{} yields sparse latents with few non-zero entries, enabling more efficient generation.

\section{Conclusion}
We introduce VAEsselSparse, a sparse encoder–decoder that combines sparse 3D convolutions with sparse attention to learn compact, organ-level representations of high-resolution vessel-like networks. By exploiting the inherent sparsity of segmentation masks, VAEsselSparse achieves $8 \times 8 \times 8$ spatial compression while accurately maintaining connectivity. Beyond reconstruction, its latent space retains clinically meaningful structure, enabling strong performance on downstream classification tasks and serving as an effective embedding space for generative modeling, supporting realistic vessel synthesis. These results suggest that sparsity-preserving representation learning is a practical path toward scalable, clinically useful modeling of high-resolution vasculature.

% \clearpage
\begin{credits}
\subsubsection{\ackname} This work has been supported by the Helmut Horten Foundation.

\subsubsection{\discintname}
The authors declare no competing interests for this article.
\end{credits}

\bibliographystyle{splncs04}
\bibliography{references}
\end{document}